\def\year{2022}\relax
\documentclass[letterpaper]{article} 
\usepackage{aaai22}  
\usepackage{times}  
\usepackage{helvet}  
\usepackage{courier}  
\usepackage[hyphens]{url}  
\usepackage{graphicx} 
\usepackage{natbib}  
\usepackage{caption} 
\DeclareCaptionStyle{ruled}{labelfont=normalfont,labelsep=colon,strut=off} 
\usepackage{algorithm}
\usepackage{algorithmicx}
\usepackage{algpseudocode}
\usepackage{amsmath}
\usepackage{amssymb}
\usepackage{amsfonts}
\usepackage{subfigure}
\usepackage{multirow}
\usepackage{newfloat}
\usepackage{listings}
\floatstyle{ruled}
\newfloat{listing}{tb}{lst}{}
\floatname{listing}{Listing}

\title{Federated Learning for Face Recognition with Gradient Correction}

\author {
    Yifan Niu, 
    Weihong Deng
}
\affiliations {
    Beijing University of Posts and Telecommunications\\
    nyf@bupt.edu.cn, whdeng@bupt.edu.cn
}

\usepackage{bibentry}

\begin{document}

\maketitle

\begin{abstract}
With increasing appealing to privacy issues in face recognition, federated learning has emerged as one of the most prevalent approaches to study  the  unconstrained face recognition problem with private decentralized data. However,  conventional decentralized  federated  algorithm sharing whole parameters of networks among clients suffers from privacy leakage in face recognition scene. In this work, we introduce a framework, FedGC, to tackle federated learning for face recognition and guarantees higher privacy. We explore a novel idea of correcting gradients from the perspective of backward propagation and propose a softmax-based regularizer to correct gradients of class embeddings by precisely injecting a cross-client gradient term. 
Theoretically, we show that FedGC constitutes a valid loss function similar to standard softmax.
  Extensive experiments have been conducted to validate the superiority of FedGC which can match the performance of conventional centralized methods utilizing full training dataset on several popular benchmark datasets.
\end{abstract}

\section{Introduction}
Face Recognition has been the prominent biometric technique for identity authentication and has been widely applied in many areas. Recently, a variety of  data-driven  approaches using Deep Convolutional Neural Networks (DCNNs) \cite{taigman2014deepface,kim2020broadface,deng2020sub,duan2019uniformface,marriott20213d} have been proposed to improve the face identification and verification accuracy.
A large scale dataset with diverse variance is crucial for discriminative face representation learning.
Although existing datasets \cite{cao2018vggface2,guo2016ms,kemelmacher2016megaface,parkhi2015deep,wang2018devil,yi2014learning,zhu2021webface260m} were created aiming to  study the  unconstrained face recognition problem, they are still biased compared with the real world data distribution.  Considering privacy issue, we are not authorized to get access to  mass face  data in real world. Thus, it is vital to train a model with private decentralized face data  to study  the  unconstrained face recognition problem in real world scene.

Federated methods on object classification  tasks are all under a common setting  where  a shallow network  is adopted as backbone and a shared fully-connected layer is applied for final classification, which is likely  to lead to privacy leakage.
 Therefore, these methods are not applicable to face recognition. Once the private class embedding is obtained, one client's  private high-fidelity face images can be easily synthesized by other clients via optimizing random noise, such as DeepInversion \cite{yin2020dreaming}. Moreover, lots of GAN-based face generation technics are also proposed to  generate a frontal photorealethistic face image with face embeddings. On the other hand, existing federated methods are mainly focusing on shallow networks(\textit{e}.\textit{g}. 2 layer fully connected network), we found these methods may easily cause network collapsing when applied to deeper network structure on facial datasets. Thus, we rethink the federated learning problem of face recognition on privacy issues, and remodel conventional Federated Averaging algorithm (FedAvg) \cite{mcmahan2017communication} via ensuring each client holds a private fully-connected layer which not only guarantees higher privacy but also contributes to network convergence.

In general, each client commonly holds a small-scale non-IID local dataset.  When we follow the above setting, onece the $k$-th client solves the optimization problem locally, the classification task is relatively  uncomplicated and the network tends to overfit and suffers from degradation of generalization ability. It leads to a phenomenon that  \emph{class embeddings} (the parameters of last fully-connected layer) of the same client are almost orthogonal to each other, but part of class embeddings of different clients are highly similar.

To solve aforementioned problem, we should constitute a new training  strategy  to train a model with private decentralized non-IID (Non Identically and Independently Distributed) facial data. In this work, we first propose FedGC, a novel and poweful federated learning framework for face recognition, which combines local  optimization and cross client optimization injected by our proposed softmax regularizer. FedGC is a privacy-preserving federated learning framework which  guarantees that each client holds private class embeddings.  In face recognition, several variants of softmax-based objective functions \cite{deng2019arcface,deng2017marginal,simonyan2014very,sun2015deeply,taigman2014deepface,wang2018additive,wolf2011face} have been proposed in centralized methods.  Hence, we propose a softmax-based regularizer  aiming to correct gradients of local softmax and precisely introduce  cross-client gradients to ensure that cross-clients class embeddings are fully spread out and it can can be readily extended to other forms.  Additionally, we give a theoretical analysis to show that FedGC constitutes a valid loss function similar to standard softmax.  Our contributions can be summarized as follows:

\begin{itemize} 

\item  We propose a federated learning framework, FedGC, for face recognition and guarantees higher privacy.  It addresses the missing local optimization problems for face-specific softmax-based loss functions. 
\item We start from a novel perspective of back propagation to correct gradients and introduce  cross-client gradients to ensure the network updates in the direction of standard softmax.  We also give a theoretical analysis to show the effectiveness and  significance of  our method.

\item Extensive experiments and ablation studies have been conducted and demonstrate the superiority of the proposed FedGC on several popular benchmark datasets.
\end{itemize}

\section{Related Work}
{\bf Face Recognition}. Face Recognition (FR) has been the prominent biometric technique for identity authentication and has been widely applied in many areas \cite{wang2018deep}. Rcently, face reogintion has achieved a series of promising breakthrough based on deep face representation learning  
 and perfromed far beyond human. Conventional face-recognition approaches are proposed such as Gabor wavelets   \cite{liu2002gabor} and LBP \cite{ahonen2006face}. Schroff \cite{schroff2015facenet} proposed triplet loss to minimize intra-class variance and maximize inter-class variance. Various of softmax-based loss functions also emerged, such as L-Softmax \cite{liu2016large}, CosFace \cite{wang2018cosface}, SphereFace \cite{liu2017sphereface}, AM-Softmax \cite{wang2018additive}, Arcface \cite{deng2019arcface}. Circle Loss \cite{sun2020circle} proposed a flexible optimization manner via re-weighting less-optimized similarity scores. GroupFace \cite{kim2020groupface} proposed a novel face-recognition achitecture learning group-aware representations.
However, these data-driven approaches aim to learn discriminative face representations 
on the premise of having the access to full private faical statistics. Public available training databases \cite{cao2018vggface2,guo2016ms,kemelmacher2016megaface,parkhi2015deep,wang2018devil,yi2014learning} are mostly collected from
the photos of celebrities due to privacy issue, it is still biased. Furthermore, with increasing appealing to privacy issues in society, existing public face datasets may turn to illegal. 

{\bf Federated Learning}. Federated Learning (FL) is a machine learning setting where many clients collaboratively train a model under the orchestration of a central server, while keeping the training data decentralized, aims to transfer the traditional deep learning methods to a privacy-preserving way.   
 Existing works seek to improve model performance, efficiency and fairness in training and communication stage. FegAvg \cite{mcmahan2017communication} was proposed  as the basic algorithm of federated learning. FedProx \cite{li2018federated} was proposed as a generalization and re-parametrization of FedAvg with a proximal term. SCAFFOLD \cite{karimireddy2019scaffold}  controls variates to correct the 'client-drift' in local updates.  
FedAC \cite{yuan2020federated} is proposed to improve convergence speed and communication efficiency. FedAwS \cite{yu2020federated} investigated a new setting where each client has access to the positive data associated with only a single class.
However, most of them are mainly focusing on shallow networks and suffers from privacy leakage in face recognition. Recently,  there also emerged some works \cite{bai2021federated,aggarwal2021fedface} focusing on federated face recognition.

\section{Methodology}

In this section, we will first provide the formulation of the federated learning and its variant for face recognition. We start by analysing, and then illustrate how we are motivated to propose FedGC.

\subsection{Problem Formulation}
We consider a $C$ class classification problem defined over a compact space $\mathcal{X}$ and a label space $\mathcal{Y}$. Let $K$ be the number of clients, suppose the $k$-th client holds the data $\{x_{i}^{k},y_{i}^{k}\}$ which distributes over $\mathcal{S}_{k}:\mathcal{X}_{k}\times\mathcal{Y}_{k}$, and ensure the identity mutual exclusion of clients   $\mathcal{Y}_{k}\cap \mathcal{Y}_{z}=\mathcal{\varnothing}$, where $k,z\in[K],k\neq{z}$, such that $\mathcal{S}=\cup_{ k\in[K]}\mathcal{S}_{k}$. In this work, we consider the following distributed optimization model:

\begin{equation}\label{eq:1}
\min\limits_{w} F(w)\triangleq \sum_{k=1}^{K}p_{k}F_{k}(w),
\end{equation}
where $p_{k}$ is the weight of the $k$-th client. Let the $k$-th client holds $n_{k}$ training data and $\sum_{k=1}^{K}n_{k}=N$, where $N$ is total number of data samples. We define $p_{k}$ as $ \frac{n_{k}}{N}$, then we have $\sum_{k=1}^{K}p_{k}=1$.

Consider an ``embedding-based'' discriminative model, given an input data $x\in\mathcal{X}$, a neural network $G:\mathcal{X}\rightarrow\mathbb{R}^{d}$ parameterized by $\theta $ embeds the data $x$ into a $d$-dimensional vector $G(x;\theta)\in \mathbb{R}^{d}$. Finally, the logits of an input data $x$ in the $k$-th client $f_{k}(x)\in\mathbb{R}^{C_{k}}$ can be expressed as:

\begin{equation}\label{eq:2}
f_{k}(x) = W_{k}^{T}G(x;\theta),
\end{equation}
where  matrix $W_{k}\in\mathbb{R}^{d\times C_{k}}$ is the class embeddings of the $k$-th client.
Then Eq. \ref{eq:1} can be reformulated as:
\begin{equation}\label{eq:3}
\min\limits_{W,\theta} F(W,\theta)\triangleq \sum_{k=1}^{K}p_{k} \frac{1}{n_{k}}\sum_{i=1}^{n_{k}} \ell _{k}\left(f_{k}(x_{i}^{k}), y_{i}^{k}\right),
\end{equation}
where $\ell _{k}(\cdot,\cdot)$ is the loss function of the $k$-th client, $W=\left[W_{1},\cdots,W_{K}\right]^T$. To provide a more strict privacy guarantee, we modifed FedAvg \cite{mcmahan2017communication} via keeping last fully-connected layer private in each client.  We term this privacy-preserving version of FedAvg  as  Federated Averaging with Private Embedding (FedPE). In FedPE, each client only have access to its own final class embeddings and the shared backbone parameters. Note that differential privacy \cite{abadi2016deep} for federated methods can be readily employed in FedPE by adding noise to the parameters from each client to enhance security.

\begin{figure*}[t]
	\begin{center}
		\includegraphics[width=1\linewidth]{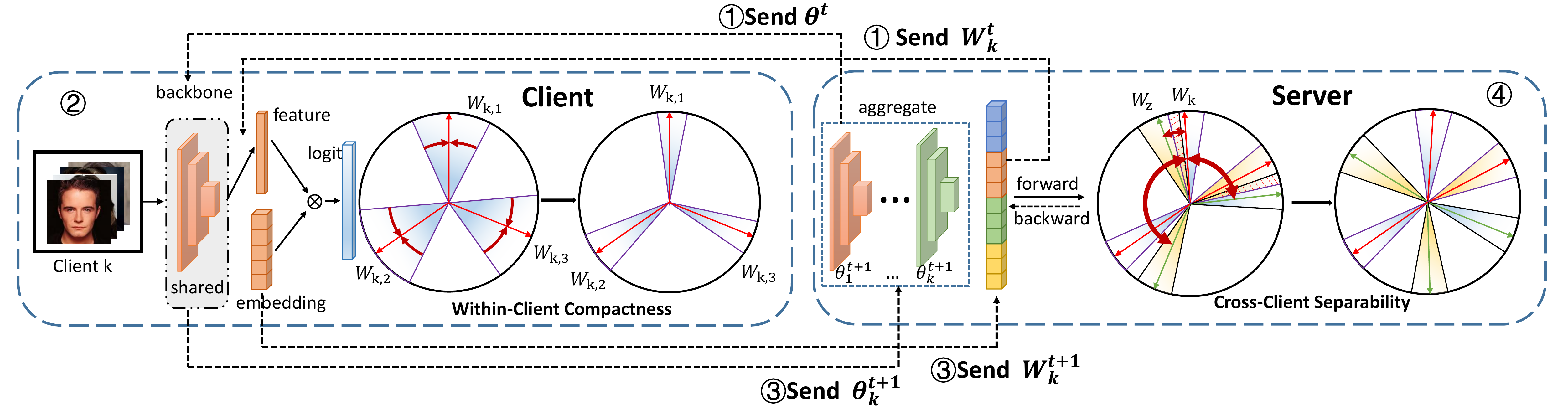}
	\end{center}
	\caption{ An illustration of our method. In communication round t, Server broadcast model parameters $\left(\theta^{t},W_{k}^{t}\right)$ to the selected clients. Then clients locally compute an update to the model  with their local data asynchronously, and send the new model $\left(\theta^{t+1},W_{k}^{t+1}\right)$ back. Finally, Server  collects an aggregate of the client updates and applys cross-client optimization. (a)Client Optimization:  clients seek to get more discriminative and more compact features. (b)Server Optimization: correct gradients and make cross-client embeddings spreadout.}
	\label{fig:2}

\end{figure*}

\subsection{Observation and Motivation}

{\bf Softmax Loss.}
Softmax loss is the most widely used calssification loss function in face recognition. For convenience, we omit the bias $b_{j}$. In the $k$-th client,  the optimization objective is to minimize the following cross-entropy loss:
\begin{equation}
L=-\frac{1}{N} \sum_{i=1}^{N} \log \frac{e^{W_{k,y_{i}}^{T} X_{k,i}}}{\sum_{j=1}^{n} e^{W_{k,j}^{T} X_{k,i}}},
\end{equation}
where $X_{k,i} \in \mathbb{R}^{d}$ denotes the deep feature of the $i$-th sample, belonging to the $y_{i}$-th class. In each individual client, the optimization objective is to minimize inter-class similarity and maximize intra-class similarity over the local class space $C_{k}$. We define this optimization in FedPE within client as  \emph{local optimization}. However, centralized training on the full training set solves the problem over the global class space $C$. We define the centralized method as \emph{global optimization}.

\begin{algorithm}[t] 
\caption{ FedGC. } 
\label{alg:1} 
\begin{algorithmic}[1]
\State {\bf Input.} The $K$ clients are indexed by $k$ and hold local data distributes over $\mathcal{S}_{k}$, $\eta$ is  learning rate.
\State Server initializes model parameters $\theta^{0}$, $W^{0}$
\For{each round $t = 0,1,...,T-1$}
\State Server initializes $k$-th client model with $\theta^{t}$, $W_{k}^{t}$.
\For{each client $k = 1,2,...,K$}
\State The $k$-th client computes local Softmax
\State $(\theta^{t+1}_{k}, W_{k}^{t+1}) \leftarrow \left(\theta^{t}, W_{k}^{t}\right)-\eta \nabla \ell _{k}\left(x_{i}^{k}, y_{i}^{k}\right)$,
\State and sends $(\theta^{t+1}_{k}, W_{k}^{t+1})$ to the server.
\EndFor
\State Server aggregates the model parameters:
\State $\theta^{t+1} \leftarrow \sum_{k=1}^{K} \frac{n_{k}}{n} \theta^{t+1}_{k}$
\State $\tilde{W}^{t+1}=\left[W_{k}^{t+1}, \ldots, W_{K}^{t+1}\right]^{T}$
\State Server applys gradient correction:
\State $W^{t+1} \leftarrow \tilde{W}^{t+1}-\lambda \eta \nabla_{\tilde{W}^{t+1}} Reg\left(\tilde{W}^{t+1}\right)$
\EndFor
\State {\bf Output.} $\theta^{T}$, $W^{T}$

\end{algorithmic} 
\end{algorithm}

In local optimization, the local softmax is to force $W_{k,y_{i}}^{T} X_{k,i}>\max _{j \in C_{k}, j \neq y_{i}}\left(W_{k,j}^{T} X_{k,i}\right)$. However, in global optimization, the softmax is to force $W_{y_{i}}^{T} X_{i}>\max _{j \in C, j \neq y_{i}}\left(W_{j}^{T} X_{i}\right)$. Thus, it is obvious that the model solving the classification problem as Eq.~\ref{eq:3} only apply within-client optimization and omit cross-client optimization,  lacking constraint $W_{k,y_{i}}^{T} X_{k,i}>\max _{j \in C_{z}, z \neq k}\left(W_{z,j}^{T} X_{k,i}\right)$. 

Therefore, this objective function as Eq.~\ref{eq:3} leads the model to a convergence state where class embeddings of the same client are almost orthogonal to each other, but part of class embeddings of different clients may highly similar. And it results in  overlapping of feature space among cross-client classes. Furthermore, only applying local optimization is more likely to cause overfitting on small-scale local datasets.

\subsection{Cross-Client Separability with Gradient Correction}
It is hard to mimic the global softmax with a set of local softmax. To address the missing optimization, as illustrated in Fig.~\ref{fig:2}, a heuristic approach to minimize similarity among cross-client class embeddings is to constrain the cross-client embeddings with a regularization  term.  Considering the additivity of gradients and the unique properties of sofmax loss gradient, we are motivated to address this issue from a new perspective of back propagation. Following the form of softmax, we define a  regularization term, namely \emph{softmax regularizer}, on the class embeddings $W\in\mathbb{R}^{d\times C}$ as:

\begin{small}
\begin{equation}\label{eq:5}
Reg\left(W\right)=\sum_{k=0}^{K} \sum_{i=0}^{C_{k}}  -\log \frac{e^{W _{k,i}^{\prime T} W_{k,i}^{\prime}}}{e^{W_{k,i}^{\prime T} W_{k,i}^{\prime}}+\sum_{z\neq k}\sum_{j=0}^{C_{z}} e^{W_{z,j}^{T} W_{k,i}^{\prime}}},
\end{equation}
\end{small}
where $W_{k,i}$ is the $i$-th class embedding of the $k$-th client, and $\left(\cdot \right)^{\prime}$ indicates the vector dosen't require gradient (the gradient is set to be zero). We precisely limit the gradient of loss function  with softmax regularizer in order to push the network update towards the direction of standard Softmax.

In addition to FedPE, the server performs an additional optimization step on the class embedding  matrix $\mathbf{W}\in\mathbb{R}^{d\times C}$ to ensure that cross-client class embeddings are separated from each other. The Federated Averaging with Gradient correction (FedGC) algorithm  is summarized in Algorithm~\ref{alg:1}.  In communication round $t$, Server broadcast model parameters $\left(\theta^{t},W_{k}^{t}\right)$ to the $k$-th clients. Then clients locally compute an update with respect to local softmax loss function with their local data asynchronously, and send the new model $\left(\theta^{t+1},W_{k}^{t+1}\right)$ back. Finally, Server  collects an aggregate of the client updates and applys cross-client optimization. Note that differential privacy can also be applied to FedGC to prevent  privacy leakage, like FedPE.

We will theoretically analyze how FedGC works and  how it pushes the network to update in the direction of global standard softmax.
Note that FedGC effectively seeks to collaboratively minimize the following objective with softmax regularizer $Reg(W)$:
\begin{equation}\label{eq:6}
F(W,\theta)\triangleq \sum_{k=1}^{K}p_{k} \frac{1}{n_{k}}\sum_{i=1}^{n_{k}} \ell _{k}\left(f_{k}(x_{i}^{k}), y_{i}^{k}\right)+ \lambda \cdot  Reg\left(W\right).
\end{equation}

For convenience, we assume that every client holds  $n_{1}=\cdots=n_{C}=\frac{N}{K}$ data and $c_{1}=\cdots=c_{K}=\frac{C}{K}$, every class holds $a_{1}=\cdots=a_{C}=\frac{N}{C}$ images, and $\lambda=\frac{1}{C}$,  the objective function  can be reformulated as:

\begin{equation}\label{eq:7}
\begin{split}
F(W,\theta) = \frac{1}{N}\sum_{k=1}^{K}\sum_{\left( x_{i},y_{i}\right)\in \mathcal{S}_{k}} \ell _{eq}\left(f_{k}(x_{i}^{k}), y_{i}^{k}\right) \\ = -\frac{1}{N}\sum_{k=1}^{K}\sum_{\left( x_{i},y_{i}\right)\in \mathcal{S}_{k}}         \left( \log \frac{e^{W_{k,y_{i}}^{T} X_{k,i}}}{\sum_{j=1}^{C} e^{W_{k,j}^{T} X_{k,i}}}  \right. \\  \left.+ \log \frac{e^{W _{k,y_{i}}^{\prime T} W_{k,y_{i}}^{\prime}}}{e^{W_{k,y_{i}}^{\prime T} W_{k,y_{i}}^{\prime}}+\sum_{z\neq k}\sum_{j=0}^{C_{z}} e^{W_{z,j}^{T} W_{k,y_{i}}^{\prime}}}\right).
\end{split}
\end{equation}

Thus, FedGC objective Eq.~\ref{eq:6} equals the empirical risk with respect to the loss function $\ell _{eq}\left(f_{k}(x_{i}^{k}), y_{i}^{k}\right)$. Our analysis easily extends to unbalanced distribution by involving a weighted  form.

Considering the collaborative effect of all the terms in $\ell _{eq}$, we give a interpretation from the perspective of backward propagation. For standard softmax in global optimization, the computation of gradients $\frac{\partial L}{\partial W_{y_{i}}}$ and $\frac{\partial L}{\partial W_{j}}$ are listed as follows:

\begin{equation}\label{eq:8}
\frac{\partial L}{\partial W_{y_{i}}}=\left(\frac{e^{W_{y_{i}}^{T} X_{i}} }{\sum_{j=1}^{C} e^{W_{j}^{T} X_{i}}}-1\right)X_{i},
\end{equation}

\begin{equation}\label{eq:9}
\frac{\partial L}{\partial W_{j}}=\frac{ e^{W_{j}^{T} X_{i}} }{\sum_{j'=1}^{C} e^{W_{j'}^{T}X_{i} }}X_{i},  where j \neq y_{i}.
\end{equation}

Similarly, for FedGC we also calculate the gradient of $\ell _{eq}$. Then, $\frac{\partial \ell _{eq}}{\partial W_{k,j}}$,  $\frac{\partial \ell _{eq}}{\partial W_{z,j}}$ and $\frac{\partial \ell _{eq}}{\partial W_{k,y_{i}}}$ can be expressed as:

\begin{equation}\label{eq:11}
\begin{split}
\frac{\partial \ell _{eq}}{\partial  W_{k,j}}=  \frac{e^{ W_{k,j}^{T} X_{k,i}} }{\sum_{j'=1}^{C_{k}} e^{W_{k,j'}^{T} X_{k,i}}}X_{k,i},
\end{split}
\end{equation}

\begin{equation}\label{eq:12}
\begin{split}
\frac{\partial \ell _{eq}}{\partial  W_{z,j}}= \frac{ e^{W_{z,j}^{T} W_{k,y_{i}}} W_{k,y_{i}}}{e^{W _{k,y_{i}}^{T} W_{k,y_{i}}}+\sum_{z\neq k}\sum_{j'=0}^{C_{z}} e^{W_{z,j'}^{T}  W_{k,y_{i}}}},
\end{split}
\end{equation}

\begin{equation}\label{eq:10}
\begin{split}
\frac{\partial \ell _{eq}}{\partial  W_{k,y_{i}}}=  \left(\frac{e^{ W_{k,y_{i}}^{T} X_{k,i}} }{\sum_{j=1}^{C_{k}} e^{W_{k,j}^{T} X_{k,i}}}-1\right)X_{k,i}.
\end{split}
\end{equation}

Let $D_{k,i}$ denote the distance between $W_{k,i}$ and $X^{k}$,  $D_{k,i} = W_{k,y_{i}}-X_{k,i}$. We assume a well trained feature on local data due to it's easy convergence on local data,  \textit{i}.\textit{e}. $D_{k,i} \to 0$, then we have $W_{k,y_{i}} \to X_{k,i} $. We can approximate:

\begin{equation}\label{eq:13}
\begin{split}
\frac{\partial \ell _{eq}}{\partial  W_{z,j}}\approx  \frac{ e^{W_{z,j}^{T} X_{k,i}}  X_{k,i}}{e^{W _{k,y_{i}}^{T}X_{k,i}}+\sum_{z\neq k}\sum_{j'=0}^{C_{z}} e^{W_{z,j'}^{T}  X_{k,i}}}.
\end{split}
\end{equation}

\begin{figure}[t]
	\begin{center}
		\includegraphics[width=1\linewidth]{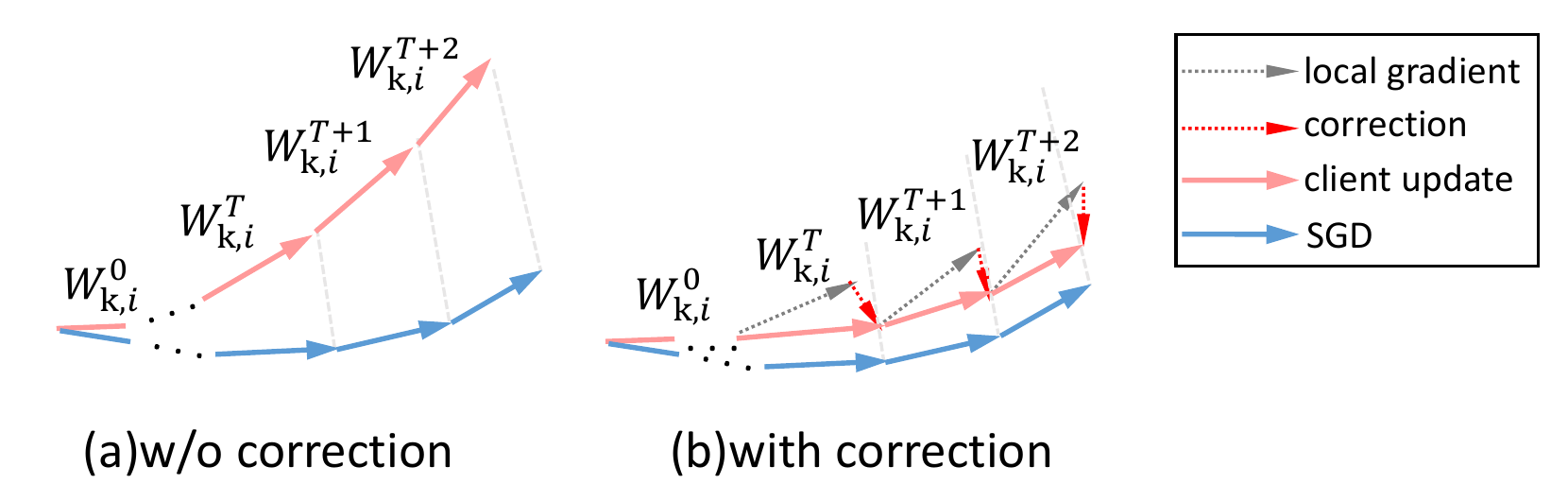}
	\end{center}
	\caption{Update steps of class embedings on a single client. (a) The divergence between FedPE and SGD becomes much larger w/o correction. (b) Gradient correction term ensures the update moves towards the true optimum.}
	\label{fig:3}
\end{figure}

The parameters are updated by SGD as $w'_{k}=w_{k}-\eta \frac{\partial \ell_{eq}}{\partial w_{k}}$,         where $\eta $ is step-size. 
Here for simplicity, we simplify Eq.~\ref{eq:8} as $\frac{\partial L}{\partial W_{y_{i}}} = \alpha X_{i}$, Eq.~\ref{eq:9}  as $\frac{\partial L}{\partial W_{j}} = \beta X_{i}$, and
Eq.~\ref{eq:10} as $\frac{\partial \ell _{eq}}{\partial  W_{k,y_{i}}} = \alpha' X_{k,i}$. Eq.~\ref{eq:11} as $\frac{\partial \ell _{eq}}{\partial  W_{k,j}} = \beta' X_{k,i}$  Eq.~\ref{eq:13} as $\frac{\partial \ell _{eq}}{\partial  W_{z,j}} = \gamma' X_{k,i}$.   We consider the direction of gradients, thus  Eq.~\ref{eq:10}  will act as a substitute for Eq.~\ref{eq:8} in within-client optimization, Eq.~\ref{eq:11} will act as a substitute for Eq.~\ref{eq:9} in within-client optimization. The collaborative effect of both terms act as local gradient in Fig.~\ref{fig:3}. The mismatch of the magitude can be alleviated by adjusting the learning rate of class embeddings.

More importantly,  Eq.~\ref{eq:13} performs cross-client optimization and act as a correction term in Fig.~\ref{fig:3} to correct gradient  in cross-client optimization,  introducing a gradient of cross-client samples to Eq.~\ref{eq:11}. And Eq.~\ref{eq:13} has the same direction as Eq.~\ref{eq:9}. And for magitude, the denominator of Eq.~\ref{eq:13} lacks term $\sum_{j=0,j\neq i}^{C_{k}} e^{W_{k,j}^{T}  X_{k,i}}$ compared to standard SGD, but with a well done local optimization, we have $\sum_{j=0,j\neq i}^{C_{k}} e^{W_{k,j}^{T}  X_{k,i}}\ll\sum_{z\neq k}\sum_{j=0}^{C_{z}} e^{W_{z,j}^{T} X_{k,i}}$. Therefore,   magitude of Eq.~\ref{eq:13} and Eq.~\ref{eq:9} are approximately equal. Thus, Eq.~\ref{eq:13} together with Eq.~\ref{eq:12} can act as a  substitute for Eq.~\ref{eq:9}, and add a missing cross-client item. Therefore,  FedGC can push the class embeddings  toward the similar dirction as standard SGD and guarantees higher privacy.


{\bf Remark:} Another simple way to introduce cross-client constraint is to minimize: $\sum_{z\neq k}\sum_{j=0}^{C_{z}} W_{z,j}^{T} W_{k,i}$, we call it cosine regularizer. For particular $W_{z,j}$,  cosine regularizer introduce gradient $\frac{\partial \ell}{\partial  W_{z,j}} = \sum_{z\neq k}W_{k,i}$. We show that our proposed softmax regularizer can act as a correction term for local softmax and also can be regarded as a weighted version of $\sum_{z\neq k}\sum_{j=0}^{C_{z}} W_{z,j}^{T} W_{k,i}$ from the perspective of backward propagation. Our proposed softmax regularizer generate gradient of larger magitude for more similar embeddings (hard example), thus it can also be  regarded as a regularization term with hard example mining.
In addition, we defined the softmax regularizer following the form of softmax.  Thus, several loss functions which are the variants of softmax (\textit{e}.\textit{g}. ArcFace,  CosFace ) can be obtained with minor modification 
on softmax regularizer. 

\subsection{Extend FedGC to More General Case}

In the above analysis, we adopt identity mutual exclusion assumption  $\mathcal{Y}_{k}\cap \mathcal{Y}_{z}=\mathcal{\varnothing}$. In fact, FedGC is to solve the problem of missing cross-client optimization.  FedGC can also be applied to  general case. We generalize the above mentioned situations, that is, some IDs are mutually exclusive and some IDs are shared. For example, there is an identity $l$ shared by a client group $K_{l}$. After each round of communication, server takes the average of $W_{n,l}^{t},  n \in K_{l}$  and apply our proposed softmax regularizer (only exclusive clients are introduced, in this case client $K-K_{l}$) to correct its gradient. In this way, we can get  $W_{l}$ updated in the direction similar to the standard softmax. With minor modifications to above analysis, we can  prove the applicability of FedGC in general case.

\subsection{Relation to Other Methods}
{\bf Multi-task learning.}  Multi-task learning combines several tasks to one system aiming to improve the generalization ability \cite{seltzer2013multi}. Considering a multi-task learning system with input data $x_{i}$, the overall objective function is a combination of several subobject loss functions, written as $L = \sum_{j}L_{j}\left(\theta,W_{j},x_{i}\right)$, where $\theta$ is generic parameters and $W_{j},j \in \left[1,2,\cdots\right]$ are task-specific parameters. While in FedGC, Eq.~\ref{eq:3} can also be regarded as a combination of many class-dependent changing tasks $L_{k}=\frac{1}{n_{k}}\sum_{j=1}^{n_{k}} \ell _{k}\left(f_{k}(x_{j}^{k}), y_{j}^{k}\right), k \in [1,\cdots,K]$.  In general, multi-task learning is conducted end-to-end and training on a single device. While in FedGC, the model is trained with class-exclusive decentralized non-IID data. Thus, our method can be also regarded as a decentralized version of multi-task learning.

{\bf Generative Adversarial Nets (GAN).} 
Based on the idea of game theory, GAN is essentially a two players minimax problem,  $\min _{G} \max _{D} V(D, G)=\mathbb{E}_{\boldsymbol{x} \sim p_{\text {data }}(\boldsymbol{x})}[\log D(\boldsymbol{x})]+\mathbb{E}_{\boldsymbol{z} \sim p_{\boldsymbol{z}}(\boldsymbol{z})}[\log (1-D(G(\boldsymbol{z})))]$,  which converges to a Nash equilibrium. In FedGC, client optimization and server optimization can be regarded as a process of adversary learning, where clients tend to minimize the similarity of within-client class embeddings,  $L_{k}=\frac{1}{n_{k}}\sum_{j=1}^{n_{k}} \ell _{k}\left(f_{k}(x_{j}^{k}), y_{j}^{k}\right)$. But server tends to  minimize the similarity of cross-client class embeddings and  encourages within-client class embeddings to be more compact,  $Reg(W)$. By performing adversary learning similar to GAN, the network can learn  more discriminative representations of class embeddings.
\section{Experiments}

\subsection{Implementation Details}

{\bf Datasets.} Considering that federated learning is extremely time-consuming, we employ CASIA-WebFace \cite{yi2014learning} as training set. CASIA-WebFace is collected from internet and contains about 1,000 subjects and 500,000 images. To simulate federated learning setting, we randomly divide training set into 36 clients. For  test, we explore the verification performance of proposed FedGC on  benchmark datasets ( LFW \cite{huang2008labeled}, CFP-FP \cite{sengupta2016frontal}, AgeDB-30 \cite{moschoglou2017agedb}, SLLFW \cite{deng2017fine}, CPLFW \cite{zheng2018cross}, CALFW \cite{zheng2017cross}, and VGG2-FP \cite{cao2018vggface2}). We also explore  on large-scale imagedatasets ( MegaFace \cite{kemelmacher2016megaface}, IJB-B \cite{whitelam2017iarpa} and IJB-C \cite{maze2018iarpa}).

{\bf Experimental Settings}. In data preprocessing, we use five facial landmarks for similarity transformation, then crop and resize the faces to (112×112). We employ the ResNet-34 \cite{he2016deep} as backbone.  We train the model with 2 synchronized 1080Ti GPUs on Pytorch.
The learning rate is set to a constant of 0.1. The learning rate is kept constant without decay which is similar to the recent federated works. The batch size is set as 256. For fair comparison, the learning rate
is also kept 0.1 in centralized standard SGD.  We set momentum as 0.9 and weight decay as 5e-4.
\subsection{Ablation Study}

{\bf Fraction of participants.} We compare the fraction of participants $C \in [0,1]$. In each communication round, there are $C\cdot K$ clients conduct optimization in parallel on LFW. Table~\ref{table:1} shows the impact of varying C for face recognition models. We train the models with  the guidance of Softmax. It is shown that with the increasing of client participation $C$, the performance of the model also increased. And FedGC still ourperforms the baseline model by a notable margin.

{\bf Regularizer multiplier $\lambda$.}
We perform an analysis of the learning rate multiplier of the softmax regularizer $\lambda$ on LFW. As shown in Table~\ref{table:2}, FedGC achieves the best performance when $\lambda$ is 20. It is shown that a large multiplier  also cause network collapsing, as it makes within-client class embeddings collapse to one point. When $\lambda$ is very small, then the model degenerates into baseline model FedPE.

{\bf Balanced \textit{v}.\textit{s}. Unbalanced Partition.} We compare the verification performance according to the partition of datasets.  Here we constructed a unbalanced partition by logarithmic normal distribution: $\ln X \sim N\left(0, 1\right)$. We perform an analysis on the model with softmax loss functions on LFW. In table~\ref{table:3}, it shows that unbalanced partition even improve the performance of network to some extent. We find that the clients which holds larger scale dataset than average contribute significantly to network and make it generate more discriminative representations.   And FedGC still outperforms baseline model on both balanced and unbalanced datasets.

\begin{table}[t]
	\begin{center}
		\resizebox{0.45\textwidth}{!}{
			\begin{tabular}{l|c|c|c|c}
\hline
Method  &$C =0.25$  & $C =0.5$ & $C =0.75$& $C =1$\\
				\hline
					Softmax-FedPE &93.12&93.83&94.32&94.77\\
				\hline
					Softmax-FedGC &{\bf 97.07}&{\bf 97.98}&{\bf 98.13}&{\bf 98.40}\\
				\hline

		\end{tabular}}
	\end{center}
	\caption{Verification performance on LFW of different  participance fraction $C$ with softmax loss function. }
	\label{table:1}
\end{table}

\begin{table}[t]
	\begin{center}
		\resizebox{0.42\textwidth}{!}{
			\begin{tabular}{l|c|c|c}
\hline
Method  &$\lambda =1$  & $\lambda =20$ & $\lambda =50$\\

				\hline
					Softmax-FedGC &95.20&98.40&97.42\\
				\hline

		\end{tabular}}
	\end{center}
	\caption{Verification performance on LFW of different learning rate multiplier  $\lambda$ with softmax loss function. }
	\label{table:2}
\end{table}
\begin{table}[t]
	\begin{center}
		\resizebox{0.45\textwidth}{!}{
			\begin{tabular}{l|c|c|c}
\hline
Method  &LFW  & CFP-FP & AgeDB\\

				\hline
				
					Balanced-FedPE &94.77&81.90&78.38\\
					Balanced-FedGC &{\bf 98.40}&{\bf 90.20}&{\bf 85.85}\\
				\hline
					Unbalanced-FedPE &96.27&85.26&81.22\\
					Unbalanced-FedGC &{\bf 98.80}&{\bf 91.56}&{\bf 88.78}\\
				\hline
		\end{tabular}}
	\end{center}
	\caption{Verification performance on LFW of different data partition with softmax loss function. }
	\label{table:3}
\end{table}
\begin{table}[t]
	\begin{center}
		\resizebox{0.35\textwidth}{!}{
			\begin{tabular}{l|c|c|c}
\hline
  &FedPE & FedCos & FedGC\\

				\hline
					LFW &94.77&96.63&{\bf 98.40}\\
				\hline

		\end{tabular}}
	\end{center}
	\caption{Verification performance on LFW of different  form  of regularization  with softmax loss function. }
	\label{table:4}
\end{table}

\begin{table*}[t]
	\begin{center}
		\resizebox{0.85\textwidth}{!}{
			\begin{tabular}{l|ccccccc|c} 
				\hline
				
					Method   &LFW  & CFP-FP & AgeDB &CALFW&CPLFW&SLLFW&VGG2-FP&Average\\
				\hline
\hline
				  	Softmax$^{*}$ &99.84 & 89.39&87.62 & 84.83& 76.08& 92.33&88.18&88.32\\
~~~~~~~~~~-FedPE  &94.77 &81.90 & 78.38& 74.15&64.40 &80.42 &80.32&79.19\\
					~~~~~~~~~~-FedPE+Fixed  &96.11 &83.67 & 80.28& 77.95&66.27 &84.23 &82.70&81.60\\
					~~~~~~~~~~-FedGC &{\bf 98.40} & {\bf 90.20}&{\bf 85.85} &{\bf 81.47} & {\bf 71.88}& {\bf 90.38}&{\bf 87.64}&{\bf 86.55}\\
				\hline
					CosFace($m=0.35$)$^{*}$  &99.10 &90.79 &91.37 &89.53 &80.20 &95.95 &89.10&90.86\\
~~~~~~~~~~-FedPE  &98.17 &86.90 & 86.28& 83.68&72.67 &91.15 &85.24&86.30\\
					~~~~~~~~~~-FedPE+Fixed &96.35 &73.01 &81.77 &79.25 &62.15 &86.57 &75.16&79.18\\
					~~~~~~~~~~-FedGC &{\bf 98.83} &{\bf 88.60} &{\bf 90.00} &{\bf 87.82} &{\bf 76.72} &{\bf 94.02} &{\bf 85.74}&{\bf 88.82}\\

				\hline
					ArcFace($m=0.5$)$^{*}$  &97.62 & 90.50&83.37 &77.33 &70.95 &86.28 &89.40&85.06\\
~~~~~~~~~~-FedPE  &98.18&87.23 &86.13 &82.47 &71.77 &91.05 &85.70&86.08\\
					~~~~~~~~~~-FedPE+Fixed & 95.85& 64.43&79.15 &77.53 &58.63 &85.67 &66.70&75.42\\
					~~~~~~~~~~-FedGC & {\bf 98.65}&{\bf 87.77} &{\bf 89.27} &{\bf 86.47} &{\bf 75.17} &{\bf 93.58} & {\bf 84.80}&{\bf 87.96}\\
\hline
				
		\end{tabular}}
	\end{center}
	\caption{ Verification results (\%) of different loss functions (Softmax, Cosface, Arcface) and  method on 7 verification datasets. FedGC  surpass others and enhance the average accuracy. $^{*}$ indicates the re-implementation by our code and $\eta $ is constant 0.1.}
	\label{table:5}
\end{table*}

{\bf Regularizer \textit{v}.\textit{s}.  Fixed.} It has been proved that random initialized vectors in high dimensional spaces (512 in this case) are almost orthogonal to each other. A na\"ive way to prevent the class embeddings from collapsing into a overlapping space is keep the class embeddings fixed to initialization.
 Table~\ref{table:5} shows that proposed FedGC outperforms model with fully-connected layer fixed ("-Fixed").    For softmax loss function, simply fixing the last fully-connected layer leads to a better accuracy. However, for Arcface  and Cosface  which introduce a more strict constraint, the performance of the model is even worse than baseline model. Intuitively, random initialized orthogonal vectors lack semantic information, and it confuses the network in a more difficult classification task. Thus, it is shown that the performance is further increased with adaptive optimization (FedGC).

{\bf Cosine \textit{v}.\textit{s}. Softmax Regularizer.}  We replace softmax regularizer with cosine regularizer, namely FedCos: $\sum_{z\neq k}\sum_{j=0}^{C_{z}} W_{z,j}^{T} W_{k,i}$, and guided by softmax loss function. We show the verification result on LFW in Table~\ref{table:4}. Although cosine regularizer shows a better accuracy than FedPE, it is still worse than FedGC. Because softmax regularizer can be regarded as a hard sample mining version of cosine regularizer, and also match the gradient in standard softmax. Thus, the superiority of softmax regularizer is proved  experimentally.

\subsection{Visualization}
\begin{figure}[t]
\centering
\subfigure[FedPE]{
\begin{minipage}[t]{0.45\linewidth}
\centering
\includegraphics[width=\linewidth]{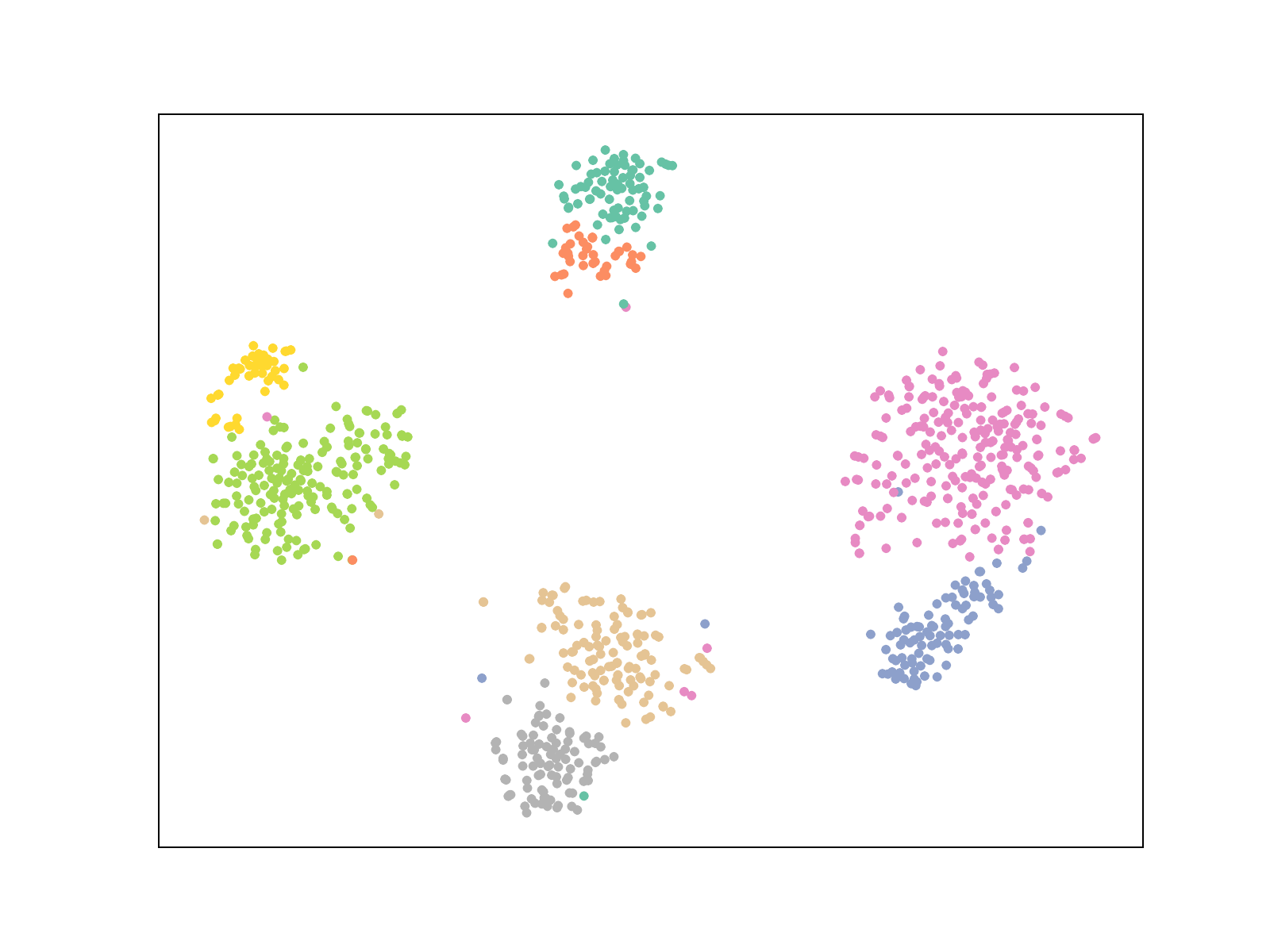}
\label{fig:4.1}
\end{minipage}%
}%
~~
\subfigure[FedGC]{
\begin{minipage}[t]{0.45\linewidth}
\centering
\includegraphics[width= 1\linewidth]{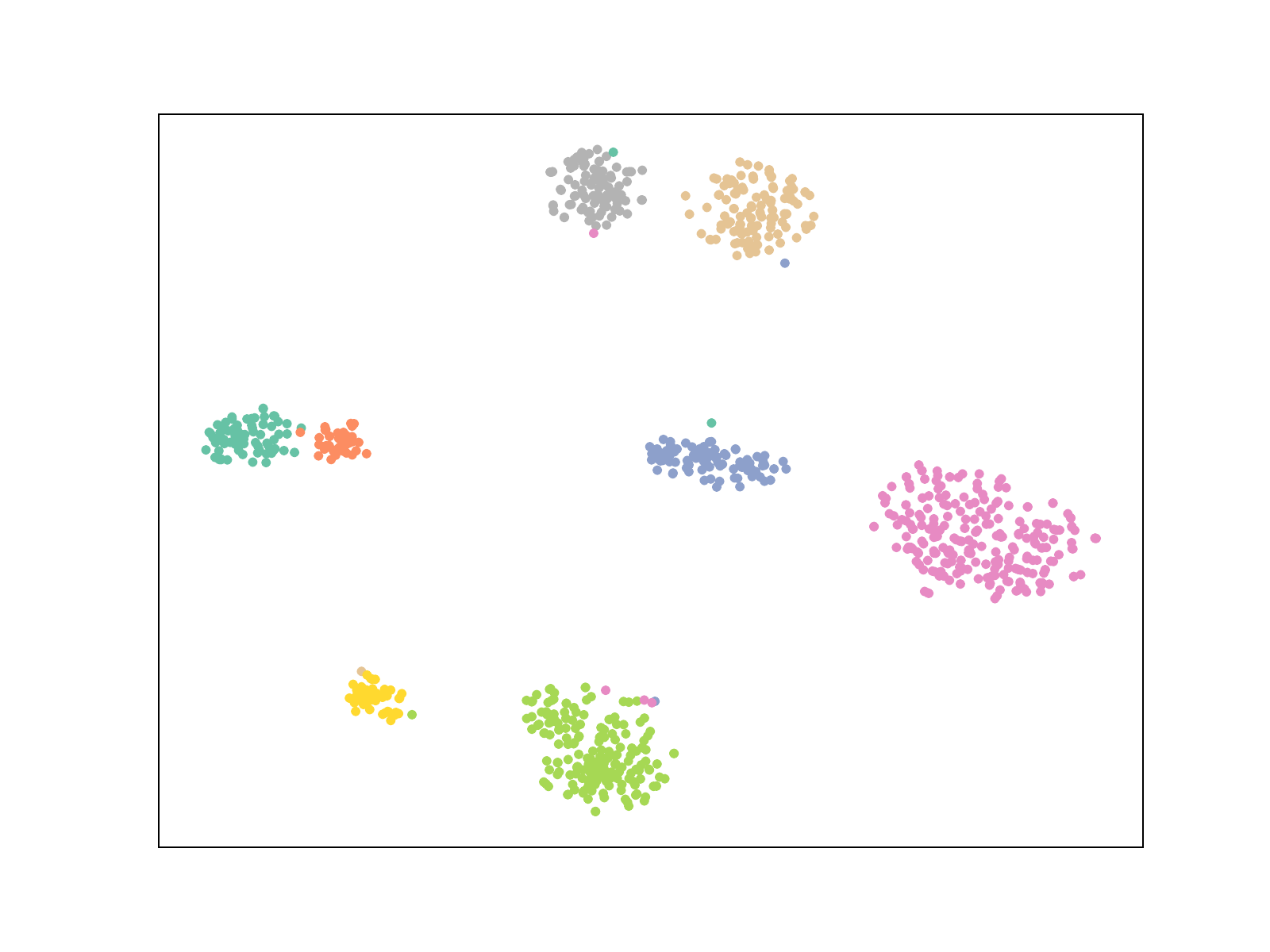}
\label{fig:4.2}
\end{minipage}%
}%

\subfigure[FedPE]{
\begin{minipage}[t]{0.45\linewidth}
\centering
\includegraphics[width=1\linewidth]{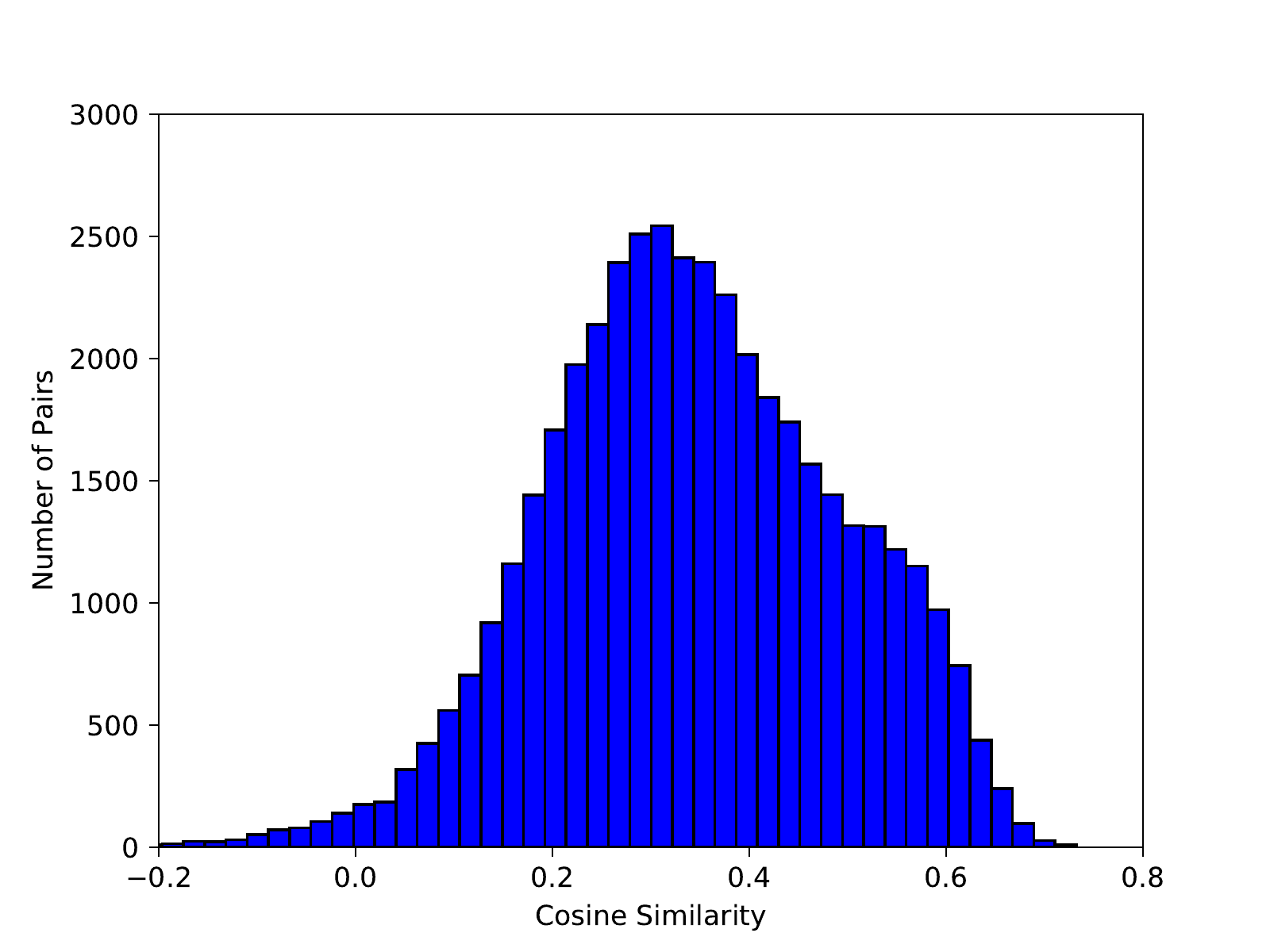}
\label{fig:4.3}
\end{minipage}%
}%
~~
\subfigure[FedGC]{
\begin{minipage}[t]{0.45\linewidth}
\centering
\includegraphics[width= 1\linewidth]{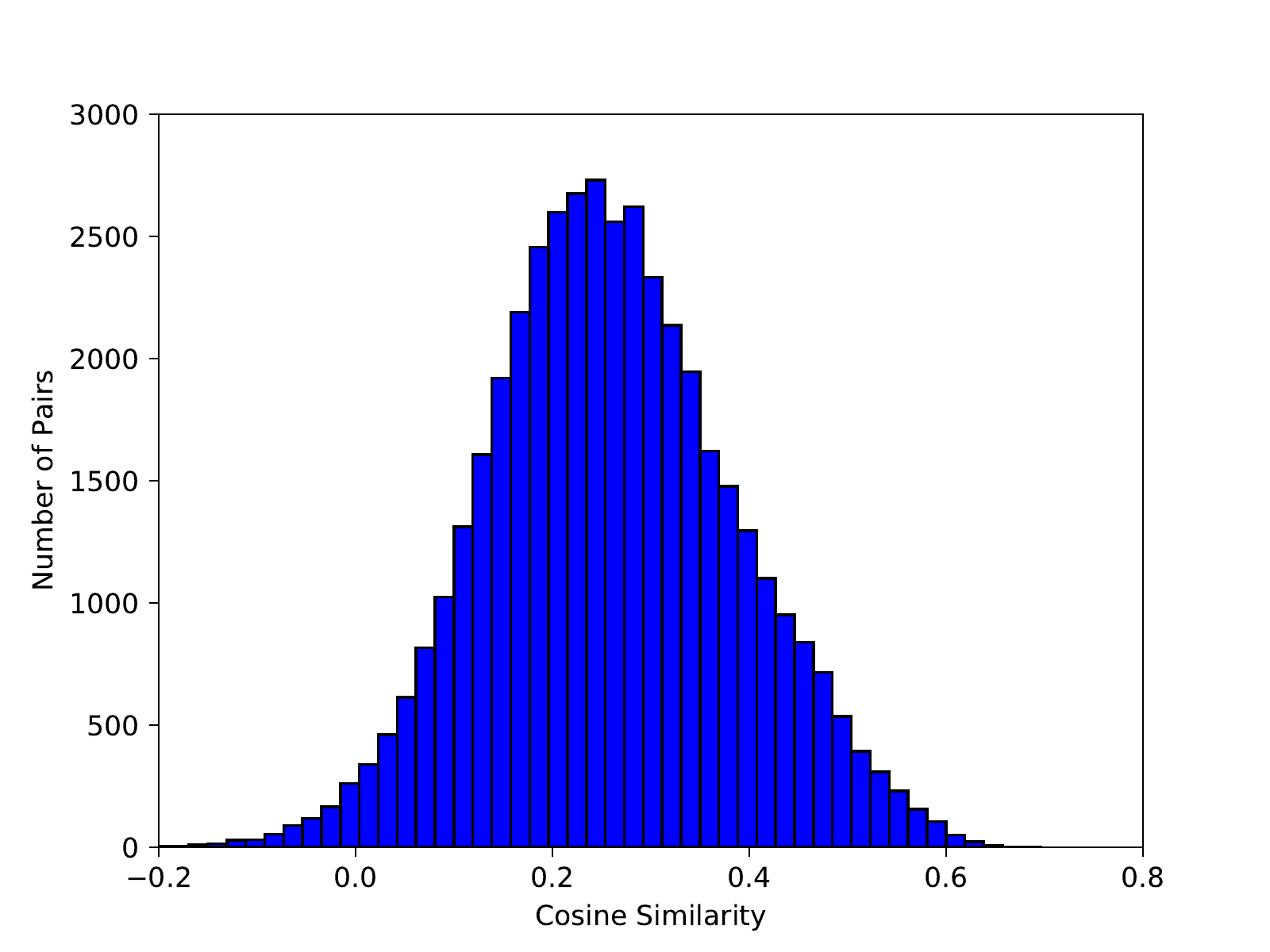}
\label{fig:4.4}
\end{minipage}%
}%
\caption{ Visualization of selected 8
classes from training set. (a)(b) t-SNE \cite{maaten2008visualizing} data distribution;
(c)(d) Histogram of pairwise cosine similarity.}

\label{fig:4}
\end{figure}

To show the effectiveness of FedGC, the visualization comparisons are conducted at feature level. We select four pairs of classes to compare FedGC and FedPE. In each pair, the two classes are from different clients and their corresponding class embeddings are highly similar in FedPE model. The features are extracted from softmax model and visualized by t-SNE \cite{maaten2008visualizing}, as shown in Fig.~\ref{fig:4.1} and Fig.~\ref{fig:4.2}, the representations of the 4 pairs tends to gather to a point and form 4 clusters in FedPE, but the representations tends to spreadout and clustered by themselves in FedGC. We also illustrate the angle distributions of all 8 selected cross-client classes. For each pair, we  calculate pair-wise cosine similarity of two classes' samples. In  Fig.~\ref{fig:4.3} and Fig.~\ref{fig:4.4}, we can clearly find that the cross-client class similarity significantly decreases in FedGC which encourage a larger cross-client class angle.

\subsection{Evaluations}
{\bf LFW, CALFW, CPLFW, CFP-FP, VGG2-FP SLLFW and AgeDB-30.}
In this section, we explore the performance of different loss functions (Softmax, Cosface \cite{wang2018cosface}, Arcface \cite{deng2019arcface}). We set the margin of Cosface \cite{wang2018cosface} at 0.35. For Arcface \cite{deng2019arcface}, we set the feature scale $s$ to 64 and choose the angular margin $m$ at 0.5. The performance of a model trained with federated learning algorithms is inherently upper bounded by that of model trained in the centralized fashion. Table~\ref{table:5} shows the experiments results, where "-X" means the dataset is trained by method "X".
FedGC achieves the highest average accuracy for all loss functions (Softmax, Arcface , Cosface) and performs better on  all of the above datasets. For ArcFace($m=0.5$), the centralized method even achieves a poor performance worse than FedPE. And FedGC can also match the performance of conventional centralized methods.

{\bf MegaFace.} The MegaFace  dataset \cite{kemelmacher2016megaface} includes 1M images of 690K different individuals as the gallery set and 100K photos of 530 unique individuals from FaceScrub \cite{ng2014data} as the probe set. It measures TPR at 1e-6 FPR for verification and Rank-1 retrieval performance for identification. In Table~\ref{table:6}, adopting FaceScrub as probe set and using the wash list provided by DeepInsight \cite{deng2019arcface},
FedGC  outperforms the baseline model FedPE by a large margin in different loss functions on both verification and identification tasks. Some centralized methods (Softmax, ArcFace($m=0.5$)) even show a poor performance when the learning rate is 0.1.  It shows that FedGC can match the performance of conventional centralized methods.

{\bf IJB-B and IJB-C.} The IJB-B dataset \cite{whitelam2017iarpa}  contains 1, 845 subjects with 21.8K still images and 55K
frames from 7, 011 videos. In total, there are 12, 115 templates with 10, 270 genuine matches and 8M impostor matches. The IJB-C dataset  \cite{maze2018iarpa}  is a further extension of IJB-B, having 3, 531 subjects with 31.3K still images and 117.5K frames from 11, 779 videos. In total, there are 23, 124 templates with 19, 557 genuine matches and 15, 639K impostor matches. The verification TPR at 1e-3 FPR and identification Rank-1 are reported in Table~\ref{table:7}. FedGC shows significant improvements and surpasses all candidates by a large margin. Compared with centralized method on all of three loss functions, FedGC can match the performance of conventional centralized methods on both IJB-B and IJB-C datasets.

\begin{table}[t]
	\begin{center}
		\resizebox{0.35\textwidth}{!}{
			\begin{tabular}{l|c|c}
\hline
Method  & Ver.(\%) &Id.(\%) \\
\hline

				  	Softmax$^{*}$ & 61.21&59.65 \\
					~~~~~~~~~~-FedPE & 36.83 &34.08 \\
					~~~~~~~~~~-FedGC & {\bf 69.87}&{\bf 61.26}\\
				\hline
					CosFace($m=0.35$)$^{*}$   & 83.30& 79.09\\
					~~~~~~~~~~-FedPE &62.62 &57.91 \\
					~~~~~~~~~~-FedGC &{\bf 72.82} &{\bf 70.96} \\

				\hline
					ArcFace($m=0.5$)$^{*}$  &50.51 & 35.18\\
					~~~~~~~~~~-FedPE & 64.53&58.12 \\
					~~~~~~~~~~-FedGC &{\bf 71.96} &{\bf 68.75}\\
\hline
		\end{tabular}}
	\end{center}
	\caption{Verification TPR (@FPR=1e-6) and identification
Rank-1 on the MegaFace Challenge 1. }
	\label{table:6}
\end{table}

\begin{table}[ht]

	\begin{center}
		\resizebox{0.45\textwidth}{!}{
			\begin{tabular}{l|cc|cc} 
				\hline
								
				\multirow{2}{*}{Method}    &\multicolumn{2}{c|}{IJB-B} &\multicolumn{2}{c}{IJB-C}\\
				\cline{2-5}
								
				 & Ver.(\%) &Id.(\%) & Ver.(\%) &Id.(\%)\\ 
				\hline
				  	Softmax$^{*}$ & 72.60&74.81 &75.06&76.05\\
					~~~~~~~~~~-FedPE  & 54.33&64.44 &57.85&65.35\\
					~~~~~~~~~~-FedGC & {\bf 69.23}&{\bf 78.52}&{\bf 71.33}&{\bf 79.52}\\
				\hline
					CosFace($m=0.35$)$^{*}$  &76.79 &78.35&79.45& 79.90\\
					~~~~~~~~~~-FedPE & 74.24&78.10&77.12& 79.10\\
					~~~~~~~~~~-FedGC & {\bf 80.28}&{\bf 82.10}&{\bf 83.40}& {\bf 83.44}\\
				\hline
					ArcFace($m=0.5$)$^{*}$  &56.64 &60.14 &59.38&59.79\\
					~~~~~~~~~~-FedPE & 73.42&76.40 &75.74&76.82\\
					~~~~~~~~~~-FedGC & {\bf 75.11}&{\bf 78.33}&{\bf 78.13}&{\bf 79.28}\\
\hline
		\end{tabular}}
	\end{center}
	\caption{ Verification TPR (@FPR=1e-3) and identification
Rank-1 on the IJB-B  and IJB-C  benchmarks.}
	\label{table:7}
\end{table}

\section{Conclusion}
In this paper, we rethink the federated learning problem for face recognition on privacy issues, and introduce a novel face-recognition-specialized federated learning framework,  FedGC,  that consists of a set of local softmax and a softmax-based regularizer to effectively learn discriminative face representtations with decentralized face data. FedGC can effectively enhance the discriminative power of cross-client class embeddings and enable the network to update towards the same direction as  standard SGD. FedGC not only guarantees higher privacy but also contributes to network convergence. Extensive experiments have been conducted over popular benchmarks to validate the effectiveness of FedGC that can match the performance of centralized methods.

{\bf Acknowledgments:} This work was supported by the National Natural Science Foundation of China under Grant 61871052.
\bibliography{egbib}
\end{document}